\documentclass[] {sig-alternate}
\usepackage{algorithmic}
\usepackage{algorithm}
\usepackage{array}
\usepackage{booktabs}
\usepackage{url}
\usepackage[tight,TABTOPCAP]{subfigure}

\newcommand{\algorithmicfunctionend}{\textbf{end function}\ }
\begin{document}
\conferenceinfo{WSDM'12,} {February 8--12, 2012, Seattle, Washingtion, USA.}
\CopyrightYear{2012}
\crdata{978-1-4503-0747-5/12/02}
\clubpenalty=10000
\widowpenalty = 10000

\title{WebSets: Extracting Sets of Entities from the Web Using Unsupervised Information Extraction}

\numberofauthors{3} 
\author{
\alignauthor
Bhavana Dalvi\\
      \affaddr{School of Computer Science}\\
      \affaddr{Carnegie Mellon University}\\
      \affaddr{Pittsburgh, PA 15213}\\
      \email{bbd@cs.cmu.edu}
\alignauthor
William W. Cohen\\
      \affaddr{School of Computer Science}\\
      \affaddr{Carnegie Mellon University}\\
      \affaddr{Pittsburgh, PA 15213}\\
      \email{wcohen@cs.cmu.edu}
\alignauthor 
Jamie Callan\\
      \affaddr{School of Computer Science}\\
      \affaddr{Carnegie Mellon University}\\
      \affaddr{Pittsburgh, PA 15213}\\
      \email{callan@cs.cmu.edu}
}

\maketitle
\begin{abstract}
We describe a open-domain information extraction method for extracting
concept-instance pairs from an HTML corpus.  Most earlier approaches to
this problem rely on combining clusters of distributionally similar
terms and concept-instance pairs obtained with Hearst patterns. In
contrast, our method relies on a novel approach for clustering terms
found in HTML tables, and then assigning concept names to these
clusters using Hearst patterns. The method can be efficiently applied
to a large corpus, and experimental results on several datasets show
that our method can accurately extract large numbers of concept-instance
pairs.
\end{abstract}

\vspace{1mm}
\noindent
{\bf Categories and Subject Descriptors:} I.2.6[Artificial Intelligence]: Learning - Knowledge acquisition

\vspace{1mm}
\noindent
{\bf General Terms:} Algorithms, Experimentation.

\vspace{1mm}

\section{Introduction} \label{sect:intro}
Many NLP tasks---include summarization, co-reference resolution, and
named entity extraction---are made easier by acquiring large sets of
concept-instance pairs (such as ``goddess,Venus'' and ``US
President'',''Bill Clinton''). Ontologies that include many such pairs
(e.g., FreeBase and WordNet) exist, but are often incomplete.  Here we
consider the problem of automatically harvesting concept-instance pairs
from a large corpus of HTML tables.

Past approaches to this problem have primarily been based detecting
\emph{coordinate terms} and \emph{hyponym patterns}.  \emph{Hyponym
patterns}, sometimes called Hearst patterns \cite{Hearst:acl1992}, are
surface patterns (like ``Xs such as Y'') indicating that X,Y are a
concept-instance pair.  Two terms $i$ and $j$ are \emph{coordinate
terms} if $i$ and $j$ are instances of the same concept: for instance,
``Bill Clinton'' and ``Richard Nixon'' are coordinate terms, since
both are instances of the concept ``US President''.  Coordinate terms
are most frequently detected by clustering terms based on
distributional similarity
\cite{Lin:coling2002}---i.e., the similarity of their contexts in free
text.  Various techniques can be used to combine these coordinate-term
and hyponym information to generate additional concept-instance pairs
\cite{VanDurme:aaai2008,Snow:nips2004}.

Our approach is novel in that it relies solely on HTML tables to
detect coordinate terms.  We present a novel clustering method that
finds extremely precise coordinate-term clusters by merging table
columns that contain overlapping \emph{triplets} of instances, and show
that this clustering method outperforms k-means, while
still being scalable enough to apply to large corpora.  We also
present a new method for combining hyponym and coordinate-term
information, and show that on table-rich corpora, this method improves
on previously-published techniques \cite{VanDurme:aaai2008}, obtaining
 higher accuracy while generating nearly hundred times the number
of concept-instance pairs.  In a final set of experiments, we show
that allowing a small amount of user input for each coordinate-term
cluster can produce concept-instance pairs with accuracy in the high
90's for four different corpora.

The experiments in this paper are conducted on several different HTML
corpora and a collection of Hearst pattern \cite{Hearst:acl1992} instances that have been
extracted from ClueWeb09, all of which are made available for
future researchers.

The rest of this paper is organized as follows. Section
\ref{sect:rel_work} presents a more detailed overview of the related work.  
Section
\ref{sect:websets} describes our unsupervised IE technique, which we call
WebSets.  Section \ref{sect:expt_eval} describes the evaluation
methodology and experimental results, and section
\ref{sect:conclusion} concludes.

\section{Related Work} \label{sect:rel_work}

Information extraction from unstructured and semi-struct\-ured information sources on the Web is an active 
area of research in recent years. In this section we summarize the existing techniques 
after categorizing them, by the inputs they are using and the goals they are trying to reach.

\subsection{Exploiting Tables on the Web}

Gatterbauer et al.\ \cite{Gatterbauer:www2007} focus on extracting
tabular data from various kinds of pages that have table-like visual
representations when rendered in a browser.  Although we do not make
use of these techniques, they could be employed by WebSets to collect
more tabular information from a corpus. 

The WebTables \cite{Cafarella:vldb2008} system extracted schema
information from a huge corpus of 14.1 billion HTML tables from
Google's general-purpose web crawl. They built an attribute
correlation statistics database (AcsDB), which can be used to create
an attribute name thesaurus and a schema auto-completion system.
Unfortunately the WebTables corpora is not publically available; it
would be an interesting project to apply WebSets to a corpus of this
size.

Gupta et al.\ \cite{Gupta:vldb2009} focuses on the task of extending a
table given a few seed rows. Their technique consolidates HTML lists
relevant to the example rows to build a result table.  Similarly, SEAL
(Set Expansion for Any Language) \cite{Wang:emnlp2009} is a set
expansion system which starts with a few seed examples and extends
them using lists detected using character-based heuristics.  Unlike
these systems, WebSets does not require seed examples, but instead
extracts concept-instance pairs in an unsupervised manner from a
corpus. 

Gupta and Sarawagi \cite{Gupta:wsdm2011} consider jointly training
structured extraction models from overlapping web source (primarily in
tables), thus avoiding the need for labeled data. WebSets also depends
on content overlap across table columns and domains, but generates
concept-instance pairs instead of building an extractor.

Limaye et al.\ \cite{Limaye:vldb2010} proposed a system to use an
existing catalog and type hierarchy for annotating table columns and
cells. They also propose better indexing techniques to improve the
search engine response to table queries.  WebSets differs in that the
Hearst-pattern data it uses is noisier than a catalog or type
hierarchy.


\subsection{Information Extraction Systems}

Many systems perform semi-supervised information extraction from free
text on the web, using only a few seed examples or seed rules.  These
systems include KnowItAll \cite{Etzioni:www2004,Etzioni:ai2005}, ASIA
\cite{Wang:acl2009}, and Coupled Pattern Learning (CPL)
\cite{Carlson:wsdm2010}. Along similar lines, Parameswaran et al.\ \cite{Parameswaran:vldb2010} propose a concept extraction algorithm which can identify a canonical
form of a concept, filtering out sub-concepts or super-concepts; and
Ritter et al.\ \cite{Ritter:aaai2009} describe a scheme for filtering
concept-instance pairs, using a SVM classifier which uses as features
frequency statistics for several Hearst patterns on a large corpus.
Given a training set of text containing known concept-instance pairs,
Snow et al. \cite{Snow:nips2004} learns ``dependency path'' features, which can
further expand the set of concept-instance pairs.
WebSets is different from most of these approaches in that it builds
all sets of entities from a given corpus, and does not require seed
sets, a starting ontology, or a set of target concept names.

TextRunner \cite{Textrunner:naacl2007} is an open-domain IE system
which makes a single pass over the corpus of unstructured text and
extracts a large set of relational tuples, without requiring any human
input.  Unlike WebSets, however, it does not build coherent sets of
category or relation instances. Pantel and Ravichandran
\cite{Pantel:naacl2004} proposes a method to automatically label
distributional term clusters using Hearst-like patterns, and Van~Durme
and Pasca \cite{VanDurme:aaai2008} proposed an alternative approach
method to extract labeled classes of instances from unstructured text.
All of these approaches use only unstructured text to find coordinate
terms and assigning hypernyms, whereas WebSets uses HTML tables. As we
show in the experimental results, WebSets uses a novel method for
combining coordinate-term clusters and hypernyms data that
quantitatively improves over Van~Durme and Pasca's method for
combining coordinate-term clusters and hypernym data.


There also exist some systems that use both free-text and tabular
information.  Talukdar et al.\ \cite{Talukdar:emnlp2008} proposed a
graph random walk based semi-supervised label propagation technique
for open domain class instance extractions, which extends the system
of Van~Durme and Pasca by using table data (from WebTables) as well as
free-text distributional clusters.  However, unlike Talukdar et al.'s
system, WebSets does not require distributional clusters.  Coupled
SEAL (CSEAL) \cite{Carlson:wsdm2010} use of mutual exclusion,
containment and type checking relationships to extend SEAL, and CPL
\cite{Carlson:wsdm2010} and CSEAL are the two components of NELL
\cite{nell:system}, a multi-strategy semi-supervised learning system.
However, unlike NELL, WebSets does not require seed instances or an
ontology as input.

Shinzato and Torisawa \cite{Shinzato:naacl2004} showed that coordinate
terms can be extracted from itemizations in structured web documents.
Their method finds concept-instance pairs pertaining to a single query
list, and finds candidate hypernyms by querying the web on-the-fly to
collect documents. In contrast, WebSets processes all lists in a
corpora simultaneously, and makes no web queries.

\section{WebSets} \label{sect:websets}
In this section we describe an unsupervised information extraction technique named WebSets
which extracts concept-instance pairs from HTML tables in a given corpus.
It builds coordinate term clusters using co-occurrence in table columns and assigns 
hypernyms to these clusters using Hearst pattern data extracted form text corpus.
To build term clusters, system should first extract tables by parsing HTML pages,
then decide which tables have useful relational data. Following the hypothesis that entities 
appearing in a table column possibly belong to the same concept, each table column in this
extracted data is a candidate entity set; the system hence needs to have a mechanism to cluster
those table columns. 
Clustering the table columns will yield sets of entities, each of which potentially belongs to a 
coherent concept. These sets will become more useful if they are labeled with appropriate concept-names.
To summarize, the technique we develop needs to solve following sub-problems:

\begin{enumerate}
 \item \textit{\textbf{Table Identification:}} Extracting tables from the corpus that 
are likely to have relational data.
 \item \textit{\textbf{Entity Clustering:}} Efficiently clustering the extracted table cells to generate
 coherent sets of entities.
 \item \textit{\textbf{Hypernym Recommendation:}} Labeling each cluster with an appropriate concept-name (hyperym).
\end{enumerate}

In this section, we describe our approach, WebSets, which solves
each of the above mentioned sub-problems in an effective way.

\subsection{Table Identification} \label{subsect:table_ident}
Currently WebSets parses tables defined by $<$table$>$ tags\footnote{We found that large number of HTML pages have broken syntax. We use the HTML syntax 
cleaning software Tidy \cite{Tidy} to fix the syntax in these pages.
}. This is only a fraction of structured data available on the Web. 
Use of other techniques like Gatterbauer et al.\ \cite{Gatterbauer:www2007}  
can provide more input data to learn sets from.

Further only a small fraction of HTML tables actually contain
useful relational data(see Section \ref{sect:expt_eval}).
Remaining tables are used for formatting or rendering purposes rather than to present relational data.
To filter out useful tables, WebSets uses the following set of features:
(1) the number of rows
(2) the number of non-link columns
(3) the length of cells after removing formatting tags
(4) whether table contains other HTML tables.
The thresholds set for our experiments are explained in Section \ref{sect:expt_eval}.

\subsection{Entity Clustering}
At the end of the table identification step, we have a collection of HTML tables which 
are likely to contain relational data. Each of the table-cells is a 
candidate entity and each table-column is a candidate set of entities.
However many columns have information that is useful only within a site 
(e.g., navigational links) and many are overlapping. 
To solve this problem we use the redundancy of information on the Web.
If a set of entities co-occur in multiple table-columns across the Web,
then it is more likely to be an important entity set.
Since these tables come from different domains and are created by different authors,
they will typically only partially overlap. To cluster the entities in these columns
into coherent sets, the first essential step will be to represent this data in a way
that reveals the co-occurrence of entities across multiple table columns. In this section,
we will first discuss the data representation, and then algorithms for entity clustering.

\subsubsection{Data Representation} \label{subsect:tripletStore}
We considered two ways to represent table data.
With the ``Entity record representation'', one record per entity is 
created, and it contains information 
about which table-columns and URL domains the entity was mentioned in. 
These entity records can then be clustered, based on the overlap of 
table-columns and domains, to yield sets of entities. 
One advantage of this representation is that it is compact. Another advantage is
that the number of distinct domains an entity occurred in can be used as an 
indicator of how ``important'' it is.
A disadvantage of this representation is that if some entity-name is ambiguous (has multiple senses),
it will collapse all senses of the entity into one record.
E.g., consider a corpus which contains mentions of the entity ``Apple'' 
in two senses, ``Apple as a fruit'' and ``Apple as a company''. 
This representation will create a single record 
for the entity ``Apple'' with its occurrence as a fruit and as a company confounded. An
unsupervised IE system might find it difficult to decompose multiple senses from this single record.

To solve this problem, we propose a novel representation of the table data called ``Entity-Triplet records''. 
In this representation, there is a record for each triplet of adjacent entities instead of for individual entities. 
Each record contains information about which table-columns and URL domains the triplet occurred in.
Hence in case of an ambiguous entity like ``Apple'', its occurrences as a fruit will be separate from its 
occurrences as a company, e.g., $\{$Apple, Avocado, Banana$\}$ will be one triplet record 
and $\{$Apple, Microsoft, DELL$\}$ will be another triplet record.
Thus entities within a triplet disambiguate each other. 
Since we have a list of all domains a triplet occurred in, only those triplets which 
appear in enough domains can be considered as important.

In this way, we represent the table data as a triplet store, which contains a record
for each entity triplet. The triplet store can be built in one single pass over the corpus.
If the system considers all possible triplets that can be created from each table column, then 
number of triplet records will be quadratic in the total size of all tables. This can be a serious concern for
a web-scale dataset. Hence our system constructs onlty those triplets which are adjacent i.e. sub-sequences of 
 entities in a table-column. This ensures that number of triplet records 
are linear in total size of all tables, making the system scalable.
The system keeps track of all table-columns a triplet occurred in, 
which makes it possible to reconstruct a column by joining triplets on columnId. 
Hence this storage method does not result in any loss of information.

\begin{table}
\centering
\scalebox{0.8}{%
\begin{tabular}{|l|l|} \hline
Country & Capital City \\
\hline
India & Delhi \\
China & Beijing \\
Canada & Ottawa \\
France & Paris \\
\hline\end{tabular}}
\caption{TableId= 21, domain= ``www.dom1.com''}
\label{table:tableEx21}
\end{table}

\begin{table}
\centering
\scalebox{0.8}{%
\begin{tabular}{|l|l|} \hline
Country & Capital City \\
\hline
China & Beijing \\
Canada & Ottawa \\
France & Paris \\
England & London \\
\hline\end{tabular}}
\caption{TableId= 34, URL= ``www.dom2.com''}
\label{table:tableEx22}
\end{table}

\begin{table}
\centering
\scalebox{0.8}{%
\begin{tabular}{|l|l|l|} \hline
Entities & Tid:Cids & Domains \\
\hline
India,China,Canada & 21:1 & www.dom1.com\\
China, Canada, France & 21:1, 34:1 & www.dom1.com, www.dom2.com\\
Delhi, Beijing, Ottawa & 21:2 & www.dom1.com\\
Beijing, Ottawa, Paris  & 21:2, 34:2 & www.dom1.com, www.dom2.com\\
Canada, England, France & 34:1 &  www.dom2.com\\
London, Ottawa, Paris & 34:2 &  www.dom2.com\\
\hline\end{tabular}}
\caption{Triplet records created by WebSets}
\label{table:tripletsEx2}
\end{table}
 
Consider an example of tables containing countries and their capitals. Original tables
are shown in Table \ref{table:tableEx21}, \ref{table:tableEx22} and 
the triplet records created by WebSets are shown in 
 Table \ref{table:tripletsEx2}. Second row in
Table \ref{table:tripletsEx2}, indicates that the triplet (China, Canada, France) occurred 
in column 1 of tableId 21 and column 1 of tableId 34. Also these entities are
retrieved from webpages which reside in the domains ``www.dom1.com'' and ``www.dom2.com''. 

These triplets are canonicalized by converting entity strings
to lower case and arranging the constituent entities in alphabetical order.
The triplets are then ranked in descending order of number of domains.

We create $O(n)$ triplets from a table column of size $n$. Adding each
triplet to the Triplet Store using hashmap takes $O(1)$ time. Given 
a set of $T$ HTML tables with a total of $N$ entities in them,  
the Triplet Store can be created in $O(N)$ time. Ranking the triplets using 
any common sorting technique will take $O(N*logN)$ time. Hence the total complexity
of building the Triplet Store is $O(N*logN)$.

\subsubsection{Building entity clusters}
The next task is to cluster these triplet records into meaningful sets. 
The system does not know how many clusters are present in the underlying
dataset; and since our dataset will be constructed from a huge HTML web corpus, the clustering 
algorithm needs to be very efficient. 
Given these requirements, it can be easily seen that parametric clustering algorithms 
like K-means may not be effective due to the unknown number of clusters.  
Non-parametric algorithms like agglomerative clustering \cite{Day:jc1984} fit most of our requirements. 
The most efficient agglomerative clustering algorithm is the single-link clustering algorithm,
but even that would require computing the similarity of every pair of 
entity triplets returned. This will be very expensive for the web scale datasets we are aiming to handle. 
Hence we develop a new bottom-up clustering algorithm which is efficient 
in terms of both space and time. Experiments to compare our algorithm with 
a standard K-means clustering algorithm are described in Section \ref{sect:expt_eval}.  

\subsubsection{\textbf{Bottom-Up Clustering Algorithm}}
 The clustering algorithm (named \textit{``Bottom-Up Clusterer''}) is
described formally in Algorithm \ref{alg:unsup-clustering}.
The clusterer scans through each triplet record $t$ which has occurred 
in at least \textit{minUniqueDomain} distinct domains.
A triplet and a cluster are represented 
with the same data-structure: (1) a set of entities, 
(2) a set of columnIds in which the entities co-occurred 
and (3) a set of domains in which the entities occurred. 

The clusterer compares the overlap of triplet $t$ against each cluster $C_i$. 
The triplet $t$ is added to the first $C_i$ 
so that either of the following two cases is true: \\
(1) at least 2 entities from $t$ appear in cluster $C_i$ .\\
(2) at least 2 columnIds from $t$ appear in cluster $C_i$.\\ (i.e. $minEntityOverlap = 2$ and $minColumnOverlap = 2$)\\
In both these cases, intuitively there is a high probability that $t$ 
belongs to the same category as cluster $C_i$. 
If no such overlap is found with existing clusters, the algorithm creates a new cluster 
and initializes it with the triplet $t$. 
	
This clustering algorithm is order dependent, i.e., if the order in which 
records are processed changes, it might return a different set of clusters. 
Finding the optimal ordering of triplets is a hard problem, 
but a reasonably good ordering can be easily generated by ordering the
triplets in the descending order of number of distinct domains. 
We discard triplets that appear in less than \textit{minUniqueDomain} domains.
\begin{algorithm}
\caption{Bottom-Up Clustering Algorithm }
\label{alg:unsup-clustering}
\begin{algorithmic}[1]
\STATE \algorithmicfunction \ Bottom-Up-Clusterer($Triplet\-Store$):$Clusters$
\COMMENT{Triplet records are ordered in descending order of number of distinct domain.}
\STATE Initialize $Clusters = \phi$; $max = 0$
 \FOR{(every $t \in Triplet\-Store$ : such that\\ \ \ \ \ \ \  $|t.domains| >= $\textit{minUniqueDomain})}
	\STATE assigned = false
	\FOR{every $C_i \in Clusters$}
	   \IF{$|t.entities \cap C_i.entities| >= minEntityOverlap$ OR 
               $|t.col \cap C_i.col| >= minColumnOverlap$}
    		\STATE $C_i = C_i \cup t$ 
		\STATE $assigned = true$
		\STATE break;
   	   \ENDIF
	\ENDFOR
	\IF{not assigned}
		\STATE increment $max$
		\STATE Create new cluster {$C_{max} = t$}
		\STATE $Clusters = Clusters \cup C_{max}$   
	\ENDIF		
\ENDFOR
\STATE \algorithmicfunctionend
\end{algorithmic}
\end{algorithm}

\subsubsection{\textbf{Computational complexity}}
Suppose that our dataset has total $T$ tables. Let these tables have in total $N$ cells. 
For each triplet $t$, Algorithm \ref{alg:unsup-clustering} 
finds entity and column overlap with all existing clusters. 
This operation can be implemented efficiently by keeping two inverted indices:
(1) from each entity to all clusterIds it belongs to and (2) from each columnId 
to all clusterIds it belongs to. ClusterIds in each ``postings list ''
\footnote[4]{In the information retrieval terminology, an inverted index has a record per word that
contains the list of all document-ids the word occurred in. This list is referred to as the ``postings list''.
In this paper, a ``postings list'' refers to the list of all clusterIds that a triplet or a columnId belongs to.} 
will be kept in sorted order.

Merging $k$ sorted lists, with resultant list size of $n$ takes $O(n*logk)$ time.
Now let us compute worst case time complexity of Algorithm \ref{alg:unsup-clustering}.
To compute entity overlap of each triplet, the algorithm merges 3 postings lists with total size of $O(N)$.
This step will take $O(N)$ time. To compute TableId:ColumnId overlap of each triplet,
it merges $O(T)$ postings lists with total size of $O(N)$. This step will take $O(N*logT)$
time. Hence for each triplet, finding a clusterId to merge the triplet with takes $O(N*logT)$ time.
There are $O(N)$ triplets. So the \textit{Bottom-Up Clusterer} 
will have worst case time complexity of $O(N^2*logT)$. In practice, it is much less than this.
If $N$ is total number of table-cells in the corpus then
the total number of triplet occurrences is also $O(N)$. Hence all the 
postings list merges can be amortized to be $O(N)$. Hence amortized complexity of this 
clustering algorithm is $O(N*logT)$. Considering the time complexity
to sort the triplet store, total time complexity is $O(N*logN)$,
 which is much better than the complexity of a naive 
single-link clustering algorithm, which is $O(N^2*logN)$.

\subsection{Hypernym Recommendation} \label{subsect:hypernym}
Previous sections described how do we obtain coordinate term clusters using co-occurrence 
of terms in the table columns. In this section we label these term clusters with the help of
Hyponym Concept dataset. 

\subsubsection{Building The Hyponym-Concept Dataset}
The \textit{Hyponym Concept Dataset} is built by acquiring concept-instance pairs from 
unstructured text using Hearst patterns.
For this task we used the data extracted from the ClueWeb09 corpus \cite{CluewebDataset} by
the developers of the NELL KB \cite{nell:system}. They used heuristics to identify
and then shallow-parse approximately 2 billion sentences, and then
extracted from this all patterns of the form ``\_ $word_1$ .. $word_k$ \_''
where the `filler' $word_1$ .. $word_k$ is between one and five tokens long,
and this filler appears at least once between two base noun phrases in
the corpus.  Each filler is paired with all pairs of noun phrases that
bracket it, together with the count of the total number of times this
sequence occurred.  For instance, the filler `\_ and vacations in \_'
occurs with the pair `Holidays, Thailand' with a count of one and
`Hotels,Italy' with a count of six, indicating that the phrase ``Hotels
and vacations in Italy'' occurred six times in the corpus .  The
hyponym dataset was constructed by finding all fillers that match one
of the regular expressions in Table \ref{table:hearst}.  These correspond to a
subset of the Hearst patterns used in ASIA \cite{Wang:acl2009} together with some
``doubly anchored'' versions of these patterns \cite{Kozareva:emnlp2010}.

\begin{table}
\centering
\scalebox{0.8}{%
\begin{tabular}{|c|c|} \hline
Id & Regular expression  \\ 
\hline
1 & arg1 such as (w+ (and$|$or))? arg2  \\
2 & arg1 (w+ )? (and$|$or) other arg2  \\
3 & arg1 include (w+ (and$|$or))? arg2  \\
4 & arg1 including (w+ (and$|$or))? arg2  \\
\hline
\end{tabular}}
\caption{Regular expressions used to create Hyponym Concept Dataset \label{table:hearst}}
\end{table}

Each record in the Hyponym Concept Dataset contains an entity and all concepts it co-occurred with.
Table \ref{table:hypernym} shows an example of records in this
dataset. According to this table, entity ``USA'' appeared with the concept ``country'' 1000 
times. Similarly, ``Monkey'' appeared with two different concepts, 100 times with ``animal'' 
and 60 times with ``mammal''.

\begin{table}
\centering
\scalebox{0.8}{%
\begin{tabular}{|c|c|} \hline
Hyponym & Concepts \\
\hline
USA & country:1000 \\
India & country:200 \\
Paris & city:100, tourist\_place:50 \\
Monkey & animal:100, mammal:60 \\
Sparrow & bird:33 \\
\hline\end{tabular}}
\caption{An example of Hyponym Concept Dataset}
\label{table:hypernym}
\end{table}

\subsubsection{Assigning Hypernyms to Clusters}

Assigning a meaningful concept-name to each entity set is important for two reasons. 
It enables us to systematically evaluate entity sets; e.g., it is easier for an evaluator 
to answer the question: ``Is Boston a city?" than ``Is Boston a member of set \#37?") 
It also makes the system more useful for summarizing the data in a corpus (see Section \ref{subsect:summ_corpus}).
This section describes how WebSets recommends candidate hypernyms for each entity set 
produced by Algorithm \ref{alg:unsup-clustering}. 

For this task, we use the coordinate term clusters extracted from tables and
\textit{Hyponym Concept Dataset} extracted using Hearst patterns. 
Note that this part of the system uses information extracted from unstructured text 
from the Web, to recommend category names to sets extracted from tables on the Web.

\begin{algorithm}
\caption{Hypernym Recommendation Algorithm}
\label{alg:hyp-reco-algo}
\begin{algorithmic}[1]
\STATE \algorithmicfunction \ GenerateHypernyms
\STATE \textbf{Given}: $c$: An entity cluster generated by Algorithm \ref{alg:unsup-clustering}, \\
               $I$: Set of all entities ,\\
               $L$: Set of all labels, \\
               $H \subseteq L \times I$: Hyponym-concept dataset, \\
\STATE \textbf{Returns:} $RL_c$ : Ranked list of hypernyms for $c$.
\STATE \textbf{Algorithm:}
 	\STATE $RL_c = \phi$
	\FOR{every label $l \in L$}
	   \STATE $H_l = $ Set of entities which co-occurred with $l$ in $H$
	   \STATE $Score(l) = |H_l \cap c|$  
	   \STATE $RL_c = RL_c \ \cup <l, Score(l)>$
	\ENDFOR
	\STATE Sort $RL_c$ in descending order of $Score(l)$	
        \STATE Output $RL_c$
\STATE \algorithmicfunctionend
\end{algorithmic}
\end{algorithm}

The algorithm is formally described in Algorithm \ref{alg:hyp-reco-algo}.
For each set produced at the end of clustering, we find which entities from the set 
belong to \textit{Hyponym Concept Dataset} and collect all concepts they co-occur with.
Then these concepts are ranked by number of unique entities in the set 
it co-occurred with. This ranked list serves as hypernym recommendations for the set.
The output of this stage can be used in two ways.
We can assign the topmost hypernym in the rank list as the class label for a cluster (used in Section \ref{subsubsect:IndHypEval}).
Another possibility is to present a ranked list of hypernyms for each cluster
to a user, who can then select the one which is the best for 
a given entity cluster (refer to Section \ref{subsubsect:WholeHypEval}).

Our method scores labels differently from Van~Durme and Pasca method \cite{VanDurme:aaai2008} 
It is also different in the sense that they output a concept-instance pair $<x,y>$ only when
$<x,y>$ appears in the set of candidate concept-instance pairs, whereas
we extend the labels to the whole cluster. Hence even if some pair $<x,y>$ is not present
in the Hyponym Concept dataset, it can be produced as output.

\section{Experimental Evaluation} \label{sect:expt_eval}
In this section we first discuss the datasets we worked on. Then we evaluate each step
of our extraction process separately. These experiments assume that our method extracts a list 
of concept-instance pairs from the HTML corpus. Later part of this section discusses
how well WebSets perform end to end, as a tool to process a large HTML corpus, and build
coherent sets of entities along with labeling each of them. The datasets and evaluations 
done for these experiments are posted online at \url{http://rtw.ml.cmu.edu/wk/WebSets/wsdm_2012_online/index.html}.

\subsection{Datasets}
To test the performance of WebSets, we created several webpage data\-sets  
 that are likely to have coherent sets of entities. An evaluation can then 
be done to check whether the system extracts expected entity sets from those datasets.
Some of these datasets are created using SEAL, CSEAL and ASIA systems 
(refer to Section \ref{sect:rel_work}) that extract
information from semi-structured pages on the Web.
Each of these systems, takes a name or seed examples of a category as input 
and finds possible instances of that category using set expansion techniques. 
It queries a web search engine during this process and stores all the pages 
downloaded from the Web in a cache, so that they can be reused for 
any similar queries in the future. The cache contents generated by these systems
help us build reasonable sized datasets for our experiments. 
We expect that WebSets will find reasonable number of HTML tables in these datasets. 

\begin{enumerate}
\item \textbf{Toy\_Apple:} This is a small toy dataset created with the help of
  multiple SEAL queries with ``Apple'' as a fruit and as a company. 
It is created to demonstrate the entity disambiguation effect of using triplet records 
and to compare various clustering algorithms.
\item \textbf{Delicious\_Sports:} This dataset is a subset of DAI-Labor Delicious corpus 
\cite{DAI-labor-delicious}, created by taking only those URLs which are tagged as ``sports".
\item \textbf{Delicious\_Music:} This dataset is a subset of DAI-Labor Delicious corpus 
\cite{DAI-labor-delicious}, created by taking only those URLs which are tagged as ``music".
 \item \textbf{CSEAL\_Useful:} CSEAL is one of the methods which
suggest new category and relation instances to NELL KB. 
CSEAL mostly extracts entities out of semi-struct\-ured information on the Web. 
For each instance in the KB, we can retrieve the information 
about which methods supported the existence of the instance.
The webpages from which these methods derived the support are also recorded.
CSEAL\_Useful dataset is a collection of those HTML pages from which
CSEAL gathered information about entities in the NELL KB.
 \item \textbf{ASIA\_NELL: }This dataset is collected using hypernyms associated 
with entities in the NELL KB as queries for ASIA. 
 Examples of such hypernyms are ``City'', ``Bird'', ``Sports team'' etc. 
 \item \textbf{ASIA\_INT:} This dataset is also collected using the ASIA 
system but with another set of category names as input. These category names come 
from ``Intelligence domain''. Examples of categories in this domain are ``government types'',
 ``international organizations'', ``federal agencies'', ``religions'' etc.
 \item \textbf{Clueweb\_HPR:} This dataset is collected by randomly sampling high pagerank pages in 
the Clueweb dataset \cite{CluewebDataset}. For this purpose we used the Fusion spam scores\cite{Kamps:trec2010} 
provided by Waterloo university and used pages with spam-rank score higher than 60\%.
\end{enumerate}

\begin{table}
\centering
\scalebox{0.8}{%
\begin{tabular}{|l|r|r|} \hline
Dataset & \#HTML  & \#tables\\ 
& pages & \\
\hline
Toy\_Apple & 574 & 2.6K\\
Delicious\_Sports & 21K  & 146.3K \\
Delicious\_Music & 183K  &  643.3K \\
CSEAL\_Useful & 30K  & 322.8K \\ 
ASIA\_NELL & 112K  & 676.9K \\ 
ASIA\_INT & 121K  & 621.3K\\ 
Clueweb\_HPR & 100K  & 586.9K\\ 
\hline\end{tabular}}
\caption{Dataset Statistics \label{table:datasets}}
\end{table}
Table \ref{table:datasets} shows the number of HTML pages and tables present in each of the above mentioned datasets.
Note that Clueweb\_HPR contains random sample of the Clueweb dataset, 
and hence does not contain large number of tables like other datasets. 
Datasets derived using SEAL or ASIA are expected to have higher concentration of
semi-structured data in them. Toy\-\_Apple and both Delicious datasets are specifically
designed for certain domains, whereas CSEAL\_Useful and ASIA\_NELL are relatively heterogeneous, Clueweb\_HPR
being the most heterogeneous dataset among all. In each of the following experiments, we work on
one or more of these datasets to evaluate different aspects of WebSets.

\subsection{Evaluation of Individual Stages} \label{susect:eval_individual}
In this section we evaluate each stage of WebSets system and measure the performance.
Here we consider WebSets as a technique to generate large number of concept-instance pairs 
given a HTML corpus. 

\subsubsection{Evaluation: Table Identification} \label{susect:eval_clusters}
Table \ref{table:tableIdent} shows the statistics of table identification for each dataset.
Based on the features described in Section \ref{subsect:table_ident}, 
we filter out only those tables as useful which cross some predefined thresholds. 
These thresholds were derived from the intuitions after manually going through some samples of data, 
and are kept constant for all datasets and experiments described in this paper.
When we evaluated a random sample of recursive tables, 93\% of them were useless for our purposes,
and only 7\% tables contained relational data. Hence we decide to ignore the recursive tables.
We construct triplets of entities in a table column, hence a table should have at least 3 rows.
As we are not using any link data in our system, we consider only those columns which do not have links.
WebSets is looking for tables containing relational data, hence if a table has at least 2 non-link
columns, probability of it having relational data increases. While counting these rows and columns,
we are looking for named entities. So from each cell, all HTML tags and links are removed.
A very coarse filtering by length is then applied. We consider only those table cells which
are 2-50 characters in length. Table \ref{table:tableIdent} shows that percentage of relational tables 
increases by orders of magnitude due to this filtering step.

\begin{table}
\centering
\scalebox{0.8}{%
\begin{tabular}{|l|r|r|r|r|r|} \hline
Dataset & \#Tables & \%Rela- & \#Filtered & \%Rela- & \#Triplets\\ 
 & & -tional & tables & -tional & \\
 & &  &  & filtered & \\
\hline
Toy\_Apple &  2.6K & 50 & 762 & 75 & 15K \\
Delicious\_Sports  & 146.3K & 15 & 57.0K & 55 & 63K\\
Delicious\_Music  & 643.3K & 20 & 201.8K & 75 & 93K\\
CSEAL\_Useful  & 322.8K & 30 & 116.2K & 80 & 1148K\\ 
ASIA\_NELL  & 676.9K & 20  & 233.0K &  55 & 421K \\ 
ASIA\_INT  & 621.3K & 15 & 216.0K &  60 & 374K \\ 
Clueweb\_HPR  & 586.9K & 10 & 176.0K & 35 & 78K \\ 
\hline
\end{tabular}}
\caption{Table Identification Statistics \label{table:tableIdent}}
\end{table}

\subsubsection{Evaluation: Clustering Algorithm} \label{susect:eval_clusters}
In these set of experiments we first compare WebSets with baseline K-means clustering algorithm.
Later we see how ``Entity-Triplet record'' representation is better than ``Entity record'' representation.
 We compare the clustering algorithms in terms of commonly used clustering metrics:
cluster purity (Purity), normalized mutual information (NMI), rand index (RI) \cite{Manning:2008} and 
Fowlkes-Mallows index (FM) which are defined as follows:
\begin{itemize}
 \item \textbf{Cluster Purity:} To compute cluster purity, for each cluster the class which is most frequent
in it gets assigned. The accuracy of this assignment is then measured
by counting the number of correctly assigned documents and dividing by total number of documents.\\
$ purity(\Omega , C) = \frac{1}{N} \sum_k \max_j |\omega_k \bigcap c_j | $ where\\
$N$ is total number of documents, $\Omega = {\omega_1,\omega_2,...,\omega_K}$ \\is the set of clusters 
and $C = {c_1, c_2, ... , c_J}$ is the set of classes. We interpret $\omega_k$ as the set of documents 
in cluster $\omega_k$ and $c_j$ as the set of documents in cluster $c_j$.
 \item \textbf{Normalized Mutual Information:} High purity is easy to achieve when the number of clusters is large, hence
purity cannot be used to trade off the quality of the clustering against the number of clusters. 
NMI is a measure that allows us to make such tradeoff. \\
$NMI(\Omega,C) = \frac{I(\Omega;C)}{[H(\Omega) + H(C)]/2}$\\
where maximum likelihood estimates for mutual information and entropy are computed as follows: \\
$I(\Omega, C) = \sum_k \sum_j \frac{|\omega_k \bigcap c_j|}{N}$ , 
$H(\Omega) = - \sum_k \frac{|\omega_k|}{N} * log \frac{|\omega_k|}{N}$.
\item \textbf{Rand Index:} This metric is based on information-theoretic interpretation of clustering. Clustering
is a series of decisions, one for each of the $N(N - 1)/2$ pairs of documents in the collection. 
True label denotes the pair belongs to same class or not. 
Predicted label denotes whether the pair belongs to same cluster or not. 
This way we can count True Positive($TP$), False positive ($FP$), True Negative ($TN$) and False Negative ($FN$) 
score for the clustering. Rand Index is then defined as follows: 
$RI = \frac{TP + TN}{TP + FP + FN + TN}$.
 \item\textbf{ Fowlkes-Mallows index:} While the previous three metrics are applicable only for hard-clustering, 
this metric is defined for soft clustering of data where clusters can be overlapping
but classes are non-overlapping.
Let $n_{ij}$ be the size of 
intersection of cluster $\omega_i$ and class $c_j$, $n_{i*} = \sum_j n_{ij}$ and $n_{*j} = \sum_i n_{ij}$.
\\
$FM = (\sum_{ij} \dbinom{n_{ij}}{2}) / \sqrt{\sum_{i}\dbinom{n_{i*}}{2} \sum_{j}\dbinom{n_{*j}}{2}}$\\
The derivation of this formula can be found in Ramirez et al. \cite{FMI:2010}.
\end{itemize}

Here each triplet is considered as a document. $\Omega$ refers to
 clusters of these triplets generated by WebSets or K-means. $C$ refers to actual classes
these triplets belong to. 
To compare quality of clusters across algorithms, we manually labeled each table-column of 
Toy\_Apple and Delicious\_Sports datasets. These labels are then extended to triplets within
the table column. This experiment is not repeated 
for remaining datasets because manual labeling of the whole dataset is very expensive.  

\begin{figure*}[ht]
\centering
 \subfigure[Cluster Purity]{
   \includegraphics[scale =.3] {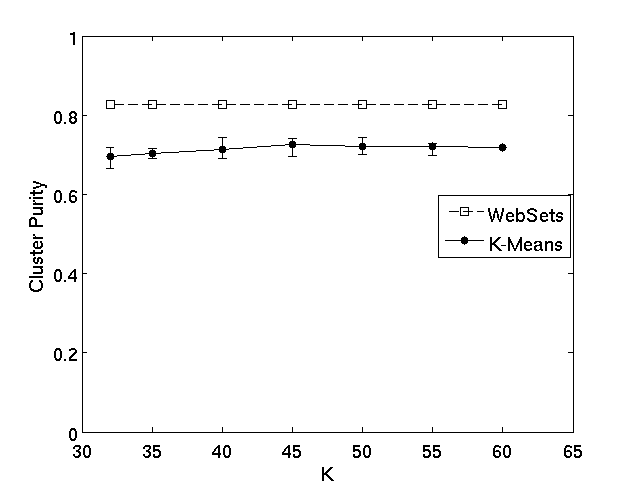}
 }
 \subfigure[Normalized Mutual Information]{
   \includegraphics[scale =.3] {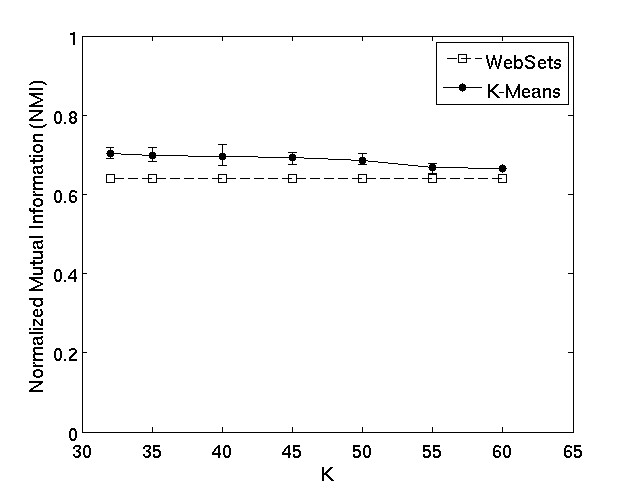}
 }
 \subfigure[Rand Index]{
   \includegraphics[scale=0.3]{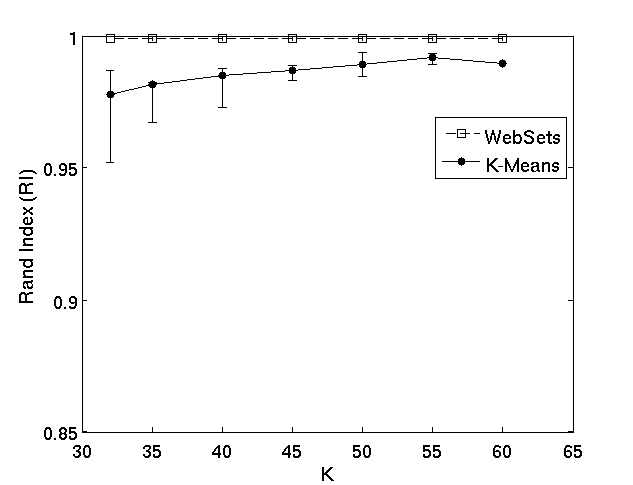}
 }
\caption{Comparison of WebSets and K-Means algorithms on Delicious\_Sports}
\label{plot:WS_vs_KMeans}
\end{figure*}

The performance of K-means depends on the input parameter $K$ and random initialization of 
cluster centroids to start the clustering process. We run K-means with cosine distance function,
 a range of values of $K$ and multiple starting points for each value of $K$. 
Figure \ref{plot:WS_vs_KMeans} shows the plot of various runs of 
K-means vs. WebSets on Delicious\_Sports dataset. 
Table \ref{table:clusterComp} shows the comparison of WebSets vs. best run of K-means on 
Toy\_Apple and Delicious\_Sports datasets.
We can see that WebSets performs better or comparable to K-means in terms of purity, NMI, RI, and FM.
Through manual labeling we found that there are 27 and 29 distinct category sets in Toy\_Apple and
Delicious\_Sports datasets respectively. We can see that WebSets defined 25 and 32 distinct clusters 
which are very close to actual number of meaningful sets, compared to 40 and 50 clusters
defined by K-means.

Standard K-means algorithm has time complexity of $O(I*K*N*T)$, where $I$ is the number of iterations,
$K$ is the number of clusters, $N$ is the number of table-cells and $T$ is the number of dimensions (here 
number of table-columns). As seen in Section \ref{sect:websets}, 
WebSets has time complexity of $O(N*log N)$. Hence our bottom-up clustering algorithm 
is more efficient than K-means in terms of time-complexity. 

To study entity disambiguation effect of triplet records, we generated both ``Entity record'' and
``Entity-Triplet record'' representations of Toy\_Apple dataset. When we ran our clustering algorithm 
on ``Entity-Triplet record'' dataset, we found ``Apple'' in two different clusters, one in which
it was clustered with other fruits and had supporting evidence from table-columns talking about fruits;
other cluster contained companies and evidence from related table-columns. Running the 
same clustering algorithm on ``Entity record'' dataset, resulted in a huge cluster containing
``Apple'', fruits and companies all combined. Thus we can say that entity triplets do help WebSets 
to disambiguate multiple senses of the same entity-string.

\vspace*{0.5mm}
\vspace*{0.5mm}


\begin{table}
\centering
\scalebox{0.9}{%
\begin{tabular}{|l|c|c|c|c|c|c|} \hline
 Dataset & Method & K & Purity & NMI & RI & FM \\ 
\hline
Toy\_Apple & K-means & 40 & 0.96 &   0.71 &   0.98 & 0.41\\
&WebSets  & 25 & 0.99 &   0.99 &   1.00	 & 0.99 \\
\hline
Delicious\_Sports &K-means & 50 & 0.72  &    0.68  &    0.98 &  0.47 \\
&WebSets & 32 & 0.83 & 0.64 &   1.00 & 	0.85 \\
\hline
\end{tabular}}
\caption{Comparison of WebSets vs. K-means\label{table:clusterComp}}
\end{table}

Now we compare the performance of clustering algorithms on entity record vs.
triplet record representation. In this experiment we use Toy\_Apple dataset. 
A table-column is considered as document and clusters produced are soft-clustering of this document set. 
Hence each table-column can be present in multiple clusters, but belongs to only one class.
Purity, NMI and RI metrics are not applicable for soft clustering, however FM metric is valid. 
We run K-Means algorithm for different values of K.
Table \ref{table:tripletVsNP} shows the best performing results of K-Means. WebSets produced 25 clusters
on entity record representation and 34 clusters using triplet representation.
In terms of FM index, each method gives better performance on triplet record representation 
when compared to entity record representation. Hence triplet record representation does improve
these clustering methods.
\begin{table}
\centering
\scalebox{0.9}{%
\begin{tabular}{|l|l|l|l|} \hline
Method & K & FM w/ Entity records & FM w/ Triplet records \\ 
\hline
WebSets &  & 0.11 (K=25)& 0.85 (K=34) \\ \hline
K-Means & 30 & 0.09 & 0.35  \\
&  25 & 0.08  & 0.38    \\
\hline
\end{tabular}}
\caption{Comparison of performance on Entity vs. Triplet record representation (Toy\_Apple dataset) \label{table:tripletVsNP}}
\end{table}

\subsubsection{Evaluation: Hyponym-Concept Dataset} \label{susect:hyponym_dataset}
Note that the Hyponym-concept dataset is created in an unsupervised way, hence we cannot guarantee that the
concept-instance pairs in this dataset are accurate. To get an idea of quality of concept-instance pairs in this dataset,
we randomly sampled 100 pairs. 55\% of them were accurate.
Hence hypernym recommendation step is dealing with noisy concept-instance pairs as input. 

\subsubsection{Evaluation: Hypernym Recommendation} \label{subsubsect:IndHypEval}
We compare our hypernym recommendation technique with Van~Durme and Pasca technique 
\cite{VanDurme:aaai2008}. Our method is different from Van~Durme and 
Pasca Method (DPM) in the sense that, 
they output a concept-instance pair only when it appears in a set of 
 candidate concept-instance pairs i.e. it exists in Hyponym-concept dataset.
In our method, based on overlap of coordinate term cluster with the Hyponym Concept dataset,
we extend the labels to whole cluster.
Another difference is the coordinate term clusters we are dealing with. DPM 
assumes term clusters are semantic partitions of terms present in the text corpus.
The clusters generated by WebSets are clusters of table columns. There is higher possibility of
having multiple small size clusters which all belong to same semantic class, but
were not merged due to our single pass bottom-up clusterer. Hence the same method may not
work equally well on these clusters.

We sample 100 concept-instance pairs randomly from output of each method to measure accuracy.
The results of different methods are presented in Table \ref{table:hypernymComp}. 
As we can see, DPM generates concept-instance pairs with 50\% accuracy even 
when run with conservative thresholds like $K=5$ and $J=0.2$. 
We also tried an extension of DPM called DPMExt which outputs a label for each entity in the cluster, 
when the label satisfies thresholds defined by $J$ and $K$. This extension increases coverage of
concept-instance pairs orders of magnitude (0.4K to 1.2K) at the cost of slight decrease 
in accuracy (50\% to 44\%).
Hypernym recommendation of WebSets (WS) is described in Algorithm \ref{alg:hyp-reco-algo}, and 
only topmost hypernym in the ranked list is produced for each cluster.
Table \ref{table:hypernymComp} shows that WS has a reasonable accuracy (62.2\%) 
and yield compared to DPM and DPMExt. As discussed in Section \ref{susect:hyponym_dataset},
Hyponym-concept dataset has noisy pairs. We also tried an extension of WebSets called
WSExt which overcomes this problem by considering only those class-instance 
pairs which have occurred at least 5 times in the corpus. Adding this simple constraint,
improves accuracy from 67\% to 78\% and correct pairs yield from 45K to 51K.

\begin{table}
\centering
\scalebox{0.85}{%
\begin{tabular}{|l|r|r|r|r|r|} 
\hline
     Method      &   K  & J & \%Accuracy &   yield (\#pairs&  \#correct pairs \\ 
                 &       &   &           &   produced)             &  (predicted) \\ \hline    
     DPM    &       $\inf$  &     0.0  &       34.6  &      88.6K   &          30.7K  \\
     DPM    &       5    &      0.2  &       50.0  &       0.8K   &         0.4K \\ \hline
     DPMExt   &   $\inf$  &     0   &   21.9  &       100,828.0K & 22,081.3K\\
     DPMExt   &   5     &     0.2  &       44.0   &      2.8K   &  1.2K \\  \hline
     WS  & - & - & 67.7 &  73.7K  &  45.8K   \\
     WSExt  & - & - & 78.8 &  64.8K  & 51.1K    \\
\hline
\end{tabular}}
\caption{ Comparison of various methods in terms of accuracy and yield on CSEAL\_Useful dataset
\label{table:hypernymComp}}
\end{table}

\subsection{WebSets as an IE technique}
In this section, we evaluate the whole WebSets system as an IE technique, which generates
coherent sets of entities and labels each set with appropriate hypernym.
At the end we demonstrate how these entity clusters can summarize the corpus.

\subsubsection{Experimental Methodology} \label{subsect:mech_turk}
 Its very expensive to label
every table-column of each dataset. Here we present a sampling based evaluation 
method to evaluate WebSets on datasets: CSEAL\_Useful, ASIA\_NELL, ASIA\_INT and Clueweb\_HPR.
We chose these four datasets to cover the different types i.e., domain-specific, 
open-domain and completely heterogeneous datasets.
There are two parts of the evaluation: evaluating labels of each cluster and 
checking whether each entity of that cluster is coherent with the assigned label.

The subjective evaluation is of the form ``deciding whether a cluster is meaningful or noisy'', 
``assigning label to an unlabeled cluster''. This is done by us (referred to as evaluators).
The objective evaluation of the kind ``whether X belongs to category Y'' is done 
using Amazon Mechanical Turk \cite{Snow:emnlp2008}.
We created yes/no questions of the form ``Is X of type Y?''. 

To evaluate precision of clusters created by WebSets, we uniformly sampled maximum 100 clusters per dataset, 
with maximum of 100 samples per cluster and gave them to the Mechanical Turk in the form 
of yes/no questions. Each question was answered by three different individuals. 
The majority vote for each question was considered as a decision for that question.  
To evaluate quality of Mechanical Turk labels, 
we sampled 100 questions at random, and manually answered the questions. 
Then we checked whether majority vote by the Mechanical Turk
matches with our answers. We specifically checked majority votes for some confusing questions 
which were likely to get labeled wrong. We found that majority vote of three individuals was 
correct more than 90\% times and in case of ambiguous questions 
precision estimates are biased low.

We evaluate WebSets using three criteria: 
(1) Are the clusters generated by WebSets meaningful? 
(2) How good are the recommended hypernyms?
(3) What is precision of the meaningful clusters? 
Subsequent sections discuss these experiments in detail.

\subsubsection{\textbf{Meaningfulness of Clusters}} \label{subsubsect:WholeHypEval}
In this experiment, we did manual evaluation of meaningfulness of clusters, 
with the help of evaluators. 
We uniformly sampled maximum 100 clusters from each dataset.
We showed following details of each cluster to the evaluator:\\
(1) top 5 hypernyms per cluster
(2) maximum 100 entities sampled uniformly from the cluster.

An evaluator was asked to look at the entities and check whether any of 
the hypernyms suggested by the system is correct. If any one of them is correct then he labels 
the cluster with that hypernym. If none of the hypernym is correct, he can label cluster
with any other hypernym that represents the cluster. If entities in a cluster are
noisy or do not form any meaningful set, then the cluster is marked as ``noisy''.
If the evaluator picks any of the candidate hypernyms as label or gives his own label 
then the cluster is considered as meaningful, else it is considered as noisy. 

Table \ref{table:meaning} shows that 63-73\% of the clusters were labeled as 
meaningful. Note that number of triplets used by the clustering algorithm 
(Table \ref{table:meaning}) is different from total number of triplets   
in the triplet store (Table \ref{table:datasets}), because
  only those triplets that occur in at least \textit{minUniqueDomain} (set to $2$ for all experiments) distinct domains are clustered. 
\begin{table}
\centering
\scalebox{0.85}{%
\begin{tabular}{|l|r|r|b{0.85in}|b{0.5in}|} \hline
 & \#Triplets & \#Clusters & \#clusters with Hypernyms& \% mean\-ingful \\ 
\hline
CSEAL\_Useful & 165.2K & 1090 & 312 & 69.0\%\\ 
ASIA\_NELL & 11.4K & 448 & 266 & 73.0\%\\ 
ASIA\_INT & 15.1K & 395 & 218 & 63.0\%\\ 
Clueweb\_HPR & 561.0 & 47 & 34 & 70.5\%\\ 
\hline\end{tabular}}
\caption{Meaningfulness of generated clusters \label{table:meaning}}
\end{table}

\subsubsection{\textbf{Performance of Hypernym Recommendation}}
In this experiment, we evaluate the performance of the hypernym recommendation using following criterion:\\
(1) What fraction of total clusters were assigned some hypernym?: This can be directly computed by 
looking at the outputs generated by hypernym recommendation. \\
(2) For what fraction of clusters evaluator chose the label from the recommended hypernyms?:
This can be computed by checking whether each of the manually assigned labels 
was one of the recommended labels.\\
(3) What is Mean Reciprocal Rank (MRR) of the hypernym ranking?: The evaluator gets to see ranked
list of top 5 labels suggested by the hypernym recommender. We compute MRR based on
rank of the label selected by the evaluator. While calculating MRR, we consider all 
meaningful clusters(including the ones for which label does not come from the recommended hypernyms).

Table \ref{table:perf_hyponym} shows the results of this evaluation.
Out of the random sample of clusters evaluated, hypernym recommendation could label 
50-60\% of them correctly. The MRR of labels is 0.56-0.59 for all the datasets. 

\begin{table}
\centering
\scalebox{0.9}{%
\begin{tabular}{|l|b{0.5in}|b{0.5in}|b{0.5in}|b{0.5in}|} \hline
 &\#Clusters Evaluated &  \#Mean\-ingful Clusters & \#Hyper\-nyms correct & MRR (meaningful)\\
\hline
CSEAL\_Useful & 100 & 69 & 57 & 0.56 \\
ASIA\_NELL & 100 & 73 & 66 & 0.59\\
ASIA\_INT & 100 & 63 & 50 & 0.58\\
Clueweb\_HPR & 34 & 24 & 20 & 0.56\\
\hline\end{tabular}}
\caption{Evaluation of Hypernym Recommender 
 \label{table:perf_hyponym}}
\end{table}

\subsubsection{\textbf{Precision of Meaningful Clusters}}
In this experiment we want to evaluate the coherency of the clusters;
i.e., whether all entities in a cluster are coherent with the label assigned to the cluster.
To verify this, we evaluated the meaningful clusters found in previous experiment, 
using the Mechanical Turk. This evaluation procedure is already discussed in Section \ref{subsect:mech_turk}.
Table \ref{table:precision_meaning} shows that the meaningful clusters\footnote[6]{First three values in column 2 of Table \ref{table:precision_meaning} are same as percentage values in column 5 of Table \ref{table:meaning}. This is because we sample maximum 100 clusters per dataset for the evaluation, hence percentage of meaningful clusters equals the actual number. For the Clueweb\_HPR dataset, there are only 47 clusters in total, so all are evaluated. } have precision 
in the range 97-99\%. This indicates that WebSets generates coherent entity clusters and hypernym assignment 
is reasonable. 

\begin{table}
\centering
\scalebox{0.8}{%
\begin{tabular}{|l|b{1in}|r|} \hline
Dataset & \#Meaningful clusters evaluated & \% Precision \\ 
\hline
CSEAL\_Useful & 69 & 98.6\%\\ 
ASIA\_NELL & 73 & 98.5\%\\ 
ASIA\_INT & 63 & 97.4\%\\ 
Clueweb\_HPR & 24 & 99.0\%\\ 
\hline\end{tabular}}
\caption{Average precision of meaningful clusters \label{table:precision_meaning}}
\end{table}

\subsubsection {Application: Summary of the Corpus} \label{subsect:summ_corpus}
The sets of entities produced by Websets can be considered as a summary of the HTML corpus
in terms of what concepts and entities are discussed in the corpus. 
Tables \ref{table:exEntDS3} and \ref{table:exEntDS6} show such summaries for the datasets 
ASIA\_INT and Delicious\_Music respectively. We choose these datasets because they are domain-specific and
hence we have some idea of what entity sets are expected to be found. 
Looking at the summary we can verify if those expectations are met. 
Due to space constraints only few clusters with ten entities from each of them are presented here. 
The labels shown here are generated by Algorithm \ref{alg:hyp-reco-algo}.

ASIA\_INT is a dataset from Intelligence domain, and the sets shown in Table \ref{table:exEntDS3} indicate that
corpus contains frequent mentions of officer ranks, international organizations, government types, religions etc.
These clusters are related to Intelligence domain and give an idea of what the corpus is about.
Similarly Table \ref{table:exEntDS3} shows the sets of entities discussed very frequently 
in Delicious\_Music i.e., music domain. 
All of them are important concepts in the music domain.
Furthermore our tool generates links to tables and domains from which this data was gathered.
While looking at these sets one can click and browse the tables and pages which mention these entities. 
Hence it can serve as a good exploratory tool for individuals
 who analyze large datasets.

\begin{table}	
\centering
\scalebox{0.85}{%
\begin{tabular}{|b{4in}|} \hline
\textbf{Religions:} Buddhism, Christianity, Islam, Sikhism, Taoism, Zoroastrianism, Jainism, Bahai, Judaism, Hinduism, Confucianism \\
\hline
\textbf{Government:} Monarchy,  Limited Democracy, Islamic Republic, Parliamentary Self Governing Territory, Parliamentary Republic, Constitutional Republic,  Republic Presidential Multiparty System, Constitutional Democracy, Democratic Republic, Parliamentary Democracy \\
\hline 
\textbf{International Organizations:} United Nations Children Fund UNICEF, Southeast European Cooperative Initiative SECI, World Trade Organization WTO, Indian Ocean Commission INOC, Economic and Social Council ECOSOC,  Caribbean Community and Common Market CARICOM, Western European Union WEU, Black Sea Economic Cooperation Zone BSEC, Nuclear Energy Agency NEA,  World Confederation of Labor WCL \\
\hline
\textbf{Languages:} Hebrew, Portuguese, Danish, Anzanian, Croatian, Phoenician, Brazilian, Surinamese, Burkinabe, Barbadian, Cuban\\
\hline\end{tabular}}
\caption{\textbf{Example Clusters from ASIA\_INT}\label{table:exEntDS3}}
\end{table}

\begin{table}
\centering
\scalebox{0.85}{%
\begin{tabular}{|b{4in}|} \hline
\textbf{Instruments:} Flute, Tuba , String Orchestra, Chimes, Harmonium, Bassoon, Woodwinds, Glockenspiel, French horn, Timpani, Piano \\
\hline
\textbf{Intervals:} Whole tone, Major sixth, Fifth, Perfect fifth, Seventh, Third, Diminished fifth, Whole step, Fourth, Minor seventh, Major third, Minor third\\
\hline 
\textbf{Genres:} Smooth jazz, Gothic, Metal rock, Rock, Pop, Hip hop, Rock n roll, Country, Folk, Punk rock\\
\hline
\textbf{Audio Equipments:} Audio editor , General midi synthesizer , Audio recorder , Multichannel digital audio workstation , Drum sequencer , Mixers , Music engraving system , Audio server , Mastering software ,  Soundfont sample player \\
\hline
\end{tabular}}
\caption{\textbf{Example Clusters from Delicious\_Music }\label{table:exEntDS6}}
\end{table}

\section{Acknowledgments}
This work is supported in part by the Intelligence Advanced Research Projects Activity
(IARPA) via Air Force Research Laboratory (AFRL) contract number
FA8650-10-C-7058. The U.S. Government is authorized to reproduce and
distribute reprints for Governmental purposes notwithstanding any
copyright annotation thereon. This work is also partially supported by the Google Research Grant.

The views and conclusions contained herein are those of the authors and should 
not be interpreted as necessarily representing the official policies or 
endorsements, either expressed or implied, of Google, IARPA, AFRL, or the U.S. Government. 

\section{Conclusion} \label{sect:conclusion}
We described a open-domain information extraction technique for extracting
concept-instance pairs from an HTML corpus.
Our approach is novel in that it relies solely on HTML tables to
detect coordinate terms.  We presented a novel clustering method that
finds extremely precise (cluster purity 83-99\%) coordinate-term clusters by merging table
columns that contain overlapping \emph{triplets} of instances. 
This clustering method outperforms k-means in terms of
Purity, Rand Index and FM index. We showed that the 
time complexity of our clustering algorithm is $O(N*logN)$, making it
more efficient than K-means or agglomerative clustering algorithms. 
We also presented a new method for combining candidate concept-instance pairs
 and coordinate-term clusters, and showed that on table-rich corpora, this method improved
on Van~Durme and Pasca method \cite{VanDurme:aaai2008}. Our method increased
the accuracy from 50\% to 78\% while generating nearly hundred times the number
of concept-instance pairs. 

We also showed that allowing a small amount of user input for labeling each coordinate-term
cluster can produce concept-instance pairs with accuracy in the range
97-99\% for four different corpora. Finally we demonstrated that 
the labeled entity sets produced by WebSets can act as summary of a HTML corpus.
The datasets and manual evaluations generated by this work will be made available for future researchers.
An interesting direction for future research can be to extend this technique to
extract the relations between the entity sets and 
naming them. 
\bibliographystyle{abbrv}
\bibliography{references}
\vspace*{0.5mm}
\scriptsize
\end{document}